\documentclass[journal]{IEEEtran}
\ifCLASSINFOpdf
\else
\fi

\usepackage{cite}
\usepackage{amsmath,amssymb,amsfonts}
\usepackage{algorithmic}
\usepackage{graphicx}
\usepackage{textcomp}
\usepackage{xcolor}

\def\b{\ensuremath\boldsymbol}

\usepackage{url}
\usepackage[algo2e,linesnumbered,lined,boxed,vlined]{algorithm2e}

\usepackage{multirow}

\begin{document}

\title{Acceleration of Large Margin Metric Learning for Nearest Neighbor Classification Using \\Triplet Mining and Stratified Sampling}

\author{Parisa Abdolrahim Poorheravi$^*$, Benyamin Ghojogh$^*$, Vincent Gaudet, Fakhri Karray, Mark Crowley
       
        
Department of Electrical and Computer Engineering, University of Waterloo, Waterloo, ON, Canada
        
Emails: \{pabdolra, bghojogh, vcgaudet, karray, mcrowley\}@uwaterloo.ca

\thanks{$^*$The first two authors contributed equally to this work.}
}


\maketitle

\begin{abstract}
Metric learning is one of the techniques in manifold learning with the goal of finding a projection subspace for increasing and decreasing the inter- and intra-class variances, respectively. Some of the metric learning methods are based on triplet learning with anchor-positive-negative triplets. Large margin metric learning for nearest neighbor classification is one of the fundamental methods to do this. Recently, Siamese networks have been introduced with the triplet loss. Many triplet mining methods have been developed for Siamese networks; however, these techniques have not been applied on the triplets of large margin metric learning for nearest neighbor classification. In this work, inspired by the mining methods for Siamese networks, we propose several triplet mining techniques for large margin metric learning. Moreover, a hierarchical approach is proposed, for acceleration and scalability of optimization, where triplets are selected by stratified sampling in hierarchical hyper-spheres. We analyze the proposed methods on three publicly available datasets, i.e., Fisher Iris, ORL faces, and MNIST datasets. 
\end{abstract}



\IEEEpeerreviewmaketitle

\section{Introduction}

\begin{figure}[!t]
\centering
\includegraphics[width=2.5in]{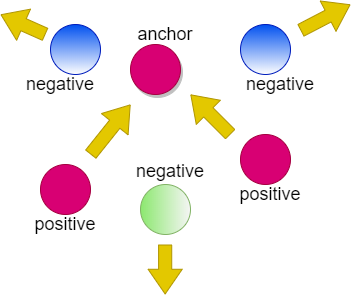}
\caption{Metric learning for decreasing and increasing the intra- and inter-class variances, respectively, by pulling the positives (same class instances) toward the anchor but pushing the negatives (other class instances) away.}
\label{figure_push_pull}
\end{figure}

Distance metric learning is one of the fundamental and most competitive techniques in machine and manifold learning \cite{kulis2013metric}. 
The goal of metric learning is to find a proper metric whose subspace discriminates the classes by increasing and decreasing the inter- and intra-class variances, respectively \cite{globerson2006metric,ghojogh2020fisher} (e.g., see Fig. \ref{figure_push_pull}). This goal was first introduced by Fisher Discriminant Analysis (FDA) \cite{globerson2006metric,friedman2001elements}. 

Some metric learning methods make use of anchor-positive-negative \textit{triplets} where the positive and negative instances are the data points having the same and different class labels with respect to an anchor instance, respectively.   
One of the first metric learning methods based on triplets was large margin metric learning for nearest neighbor classification \cite{weinberger2006distance,weinberger2009distance}. This method uses Semi-Definite Programming (SDP) optimization \cite{vandenberghe1996semidefinite} as SDP has been found to be useful for metric learning \cite{weinberger2006distance,weinberger2009distance,alipanahi2008distance,weinberger2006unsupervised}. 
Later, the concept of a triplet cost function was proposed in the field of neural networks by introducing Siamese networks \cite{hadsell2006dimensionality,schroff2015facenet,hoffer2015deep}. The triplet loss can be either in the form of Hinge loss \cite{schroff2015facenet} or softmax \cite{ye2019unsupervised}. The examples of the former and latter are \cite{weinberger2006distance,weinberger2009distance,schroff2015facenet} and \cite{goldberger2005neighbourhood,movshovitz2017no,xuan2020improved}, respectively. 

Solving SDP problems requires the interior point method \cite{boyd2004convex}, which is iterative and slow especially for big data. This can be improved and accelerated by selecting the most important data points for embedding \cite{wu2017sampling}. For example, we rather care about the nearest or farthest positives and negatives than selecting all the data points. This technique is referred to as \textit{triplet mining} in the literature where the positive and negative instances with respect to an anchor make a triplet \cite{schroff2015facenet}. 

After the introduction of Siamese networks in the literature, different triplet mining techniques were developed for Siamese training using triplets. However, these mining methods have not been implemented or proposed for the previously developed concept of large margin metric learning for nearest neighbor classification. In this work, inspired by the mining techniques for  Siamese networks, we propose different triplet mining methods for large margin metric learning. By only considering the most valuable points of the dataset with respect to anchors, the SDP optimization speeds up while preserving an acceptable classification accuracy in large margin metric learning.

In addition to proposing triplet mining techniques for the optimization, we propose a hierarchical approach for further acceleration of metric learning. This approach includes iterative selection of data subsets by hierarchical stratified sampling \cite{barnett1974elements} to train the embedding subspace. Not only does this approach accelerate the SDP optimization by reducing time complexity, but also it improves performance in some cases due to effectiveness of model averaging \cite{claeskens2008model} and reduction of estimation variance by stratified sampling \cite{ghojogh2019theory}. We also used the proposed triplet mining techniques in combination with the proposed hierarchical approach for the sake of acceleration.

The remainder of paper is as follows. In Section \ref{section_literature_review}, we review the foundations of large margin metric learning, triplet loss, and Siamese networks. 
We discuss the triplet mining methods that have already been proposed for Siamese triplet training, i.e., batch all \cite{ding2015deep}, batch hard \cite{hermans2017defense}, batch semi-hard \cite{schroff2015facenet}, easiest/hardest positives and easiest/hardest negatives, and negative sampling \cite{wu2017sampling}. 
Section \ref{section_proposed_solution} proposes how to use the triplet mining techniques in SDP optimization of large margin metric learning. The hierarchical approach is proposed in Section \ref{section_hierarchical}. We report the experimental results in Section \ref{section_experiments}. Finally, Section \ref{section_conclusion} concludes the paper and provides the possible future direction.


\section{Background}\label{section_literature_review}

\subsection{Large Margin Metric Learning for Nearest Neighbor Classification}

$k$-Nearest Neighbor ($k$-NN) classification is highly impacted by the distance metric utilized for measuring the differences between data points. Euclidean distance does not weight the points and it values them equally. A general distance metric can be viewed as the Euclidean distance after projection of points onto a discriminative subspace. This projection can be viewed as a linear transformation with a projection matrix denoted by $\b{L}$ \cite{jolliffe2011principal}. We call this general metric the Mahalanobis distance \cite{kulis2013metric,globerson2006metric}:
\begin{equation}\label{equation_metric}
\begin{aligned}
\mathcal{D} &:=\|\b{x}_i- \b{x}_j\|_{\b{M}}^2 := \|\b{L}^\top (\b{x}_i - \b{x}_j)\|_2^2 \\
&= (\b{x}_i - \b{x}_j)^\top \b{M} (\b{x}_i - \b{x}_j),
\end{aligned}
\end{equation}
where $\b{M} := \b{L} \b{L}^\top$. The matrix $\b{M}$ must be positive semi-definite, i.e. $\b{M} \succeq 0$, for the metric to satisfy convexity and the triangle inequality \cite{boyd2004convex}. 

In order to improve the $k$-NN classification performance, we should decrease and increase the intra- and inter-class variances of data, respectively \cite{ghojogh2020fisher}. As can be seen in Fig. \ref{figure_push_pull}, one way to achieve this goal is to pull the data points of the same class toward one another while pushing the points of different classes away. 

Let $y_{il}$ be one (zero) if the data points $\b{x}_i$ and $\b{x}_l$ are (are not) from the same class. Moreover, let $\eta_{ij}$ be one if $\b{x}_j$ is amongst the $k$-nearest neighbors of $\b{x}_i$ with the same class label; otherwise, it is zero. For tackling the goal of pushing together the points of a class and pulling different classes away, the following cost function can be minimized \cite{weinberger2006distance}:
\begin{equation}\label{equation_cost_1}
\begin{aligned}
&\sum_{i,j} \eta_{ij} \|\b{L}^\top (\b{x}_i - \b{x}_j)\|_2^2 + c \sum_{i,j,l} \eta_{ij} (1 - y_{il}) \Big[1 + \\
&~~~~~~~\quad\quad\qquad \|\b{L}^\top (\b{x}_i - \b{x}_j)\|_2^2 - \|\b{L}^\top (\b{x}_i - \b{x}_l)\|_2^2\Big]_+,
\end{aligned}
\end{equation}
where $[.]_+ := \max(.,0)$ is the standard Hinge loss. 
The first term in Eq. (\ref{equation_cost_1}) pushes the same-class points towards each other. The second term, on the other hand, is a triplet loss \cite{schroff2015facenet} which increases and decreases the inter- and intra-class variances, respectively. 

Inspired by support vector machines, the cost function (\ref{equation_cost_1}) can be restated using slack variables:
\begin{equation}\label{equation_cost_2}
\begin{aligned}
& \underset{\b{M},\, \xi_{ijl}}{\text{minimize}}
& & \mathcal{L} := \sum_{i,j} \eta_{ij}\, \|\b{x}_i - \b{x}_j\|^2_{\b{M}} \\
& & &~~~~~~~~~~~~~ + c \sum_{i,j} \eta_{ij}\, (1 - y_{il})\, \xi_{ijl}, \quad \forall l \\
& \text{subject to}
& & \|\b{x}_i - \b{x}_l\|^2_{\b{M}} - \|\b{x}_i - \b{x}_j\|^2_{\b{M}} \geq 1 - \xi_{ijl}, \\
& & & \xi_{ijl} \geq 0, \\
& & & \b{M} \succeq \b{0},
\end{aligned}
\end{equation}
which is a SDP problem \cite{vandenberghe1996semidefinite}. The first term in the objective functions of Eqs. (\ref{equation_cost_1}) and (\ref{equation_cost_2}) are equivalent because of Eq. (\ref{equation_metric}). The Hinge loss in Eq. (\ref{equation_cost_1}) can be approximated using non-negative slack variables, denoted by $\xi_{ijl}$. The second term of objective function in Eq. (\ref{equation_cost_2}), in addition to the first and second constraints, play the role of Hinge loss. 

\subsection{Triplet loss and Siamese Network}

As explained for Eq. (\ref{equation_cost_1}), the second term in that equation is the triplet loss which pushes the classes away and pulls the points of a class together. In Eq. (\ref{equation_cost_1}), $\b{x}_i$, $\b{x}_j$, and $\b{x}_l$ are anchor, positive, and negative instances, respectively. The goal of triplet loss is to make anchor and positive instances closer and push the negative instances away as also seen in Fig. \ref{figure_push_pull}. 

Recently, the triplet loss has been used for training neural networks which are called Siamese or triplet networks \cite{schroff2015facenet}. A Siamese network is composed of three sub-networks which share their weights. The anchor, positive, and negative instances are fed to these sub-networks and the triplet loss is used to tune their weights. Siamese networks are usually used for learning a discriminative embedding space. In this work, we propose several triplet mining methods inspired by the triplet mining techniques already existing in the literature for the Siamese nets. 

\section{Proposed Triplet Mining}\label{section_proposed_solution}

The optimization problem in Eq. (\ref{equation_cost_2}) considers all the negative instances even in large datasets. 
The SDP for solving Problem (\ref{equation_cost_2}) is very time-consuming and slow \cite{vandenberghe1996semidefinite}. Hence, Problem (\ref{equation_cost_2}) becomes intractable for large datasets, as has been noted in \cite{weinberger2006distance}. 
This motivated us to use triplet mining on the data for further improvement upon \cite{weinberger2006distance,weinberger2009distance}. 
There exist several triplet mining methods which are proposed for Siamese network training. Inspired by those, we propose here the triplet mining techniques in the objective function of Eq. (\ref{equation_cost_2}) to facilitate the optimization process. 
There can be different ways for triplet mining. In the following, we propose $k$-batch all, $k$-batch hard, $k$-batch semi-hard, extreme distances, and negative sampling for large margin metric learning.

\subsection{$k$-Batch All}

One of the mining methods to be considered is \textit{batch all} which takes all the positives and negatives of the data batch into account for Siamese neural network  \cite{ding2015deep}. 
The proposed method in \cite{weinberger2006distance,weinberger2009distance} is a batch-all version which takes only $k$ nearest positives and all the negatives. This makes sense because the SDP is slow and cannot handle all possible permutations of positive and negative instances. 
Here, we call this method \textit{$k$-batch all ($k$-BA)} where the objective in equation:
\begin{equation}
\begin{aligned}
\mathcal{L} = &\sum_{i,j} \eta_{ij}\,  \|\b{x}_i - \b{x}_j\|^2_{\b{M}}  + c \sum_{i,j} \eta_{ij}\, (1 - y_{il})\, \xi_{ijl}, \quad \forall l.
\end{aligned}
\end{equation}

\subsection{$k$-Batch Hard}\label{section_batch_hard}

Another mining method for Siamese networks is \textit{batch hard} in which the farthest positive and nearest negative with respect to the anchor are considered \cite{hermans2017defense}. The farthest positive is the hardest one to be classified as a neighbor of the anchor. Likewise, the nearest negative is the hardest one to be separated from the anchor's class. In this work, we consider $k$ positive and $k$ negative instances and we call this \textit{$k$-batch hard ($k$-BH)} where the objective in Eq. (\ref{equation_cost_2}) becomes:
\begin{equation}
\begin{aligned}
\mathcal{L} = &\sum_{i,j} \gamma_{ij}^+\, \|\b{x}_i - \b{x}_j\|^2_{\b{M}} + c \sum_{i,j,l} \gamma_{ij}^+\, \eta_{il}^-\, (1 - y_{il})\, \xi_{ijl},
\end{aligned}
\end{equation}
where $\gamma_{ij}^+$ is one (zero) if $\b{x}_j$ is (is not) amongst the $k$-farthest neighbors of $\b{x}_i$ with the same class label. Similarly, $\eta_{il}^-$ is one (zero) if $\b{x}_l$ is (is not) amongst the $k$-nearest neighbors of $\b{x}_i$ with different class label.

\subsection{$k$-Batch Semi-Hard}

\textit{Batch semi-hard} is another method, for Siamese networks, in which the hardest negatives (closest to the anchor) that are farther from the positive are taken into account \cite{schroff2015facenet}. In our work, we have $k$ positive instances and for each, we consider $k$ negatives. This method we call \textit{$k$-batch semi-hard ($k$-BSH)} in which the cost in Eq. (\ref{equation_cost_2}) can be modeled as:
\begin{equation}
\begin{aligned}
\mathcal{L} = &\sum_{i,j} \eta_{ij}\, \|\b{x}_i - \b{x}_j\|^2_{\b{M}} + c \sum_{i,j} \eta_{ij}\, \sum_{l} \eta_{il}^\mp\, (1 - y_{il})\, \xi_{ijl},
\end{aligned}
\end{equation}
where $\eta_{ij}$, as defined before, is one (zero) if $\b{x}_j$ is (is not) amongst the $k$-nearest neighbors of $\b{x}_i$ with the same class label and $\eta_{il}^\mp$ is one (zero) if $\b{x}_l$ is (is not) amongst the $k$-nearest neighbors of $\b{x}_i$, with different class label, and farther from $\b{x}_j$ to $\b{x}_i$.

\subsection{Extreme Distances}

Considering that every instance could be chosen based on their distance to the anchor (whether they are nearest or farthest), we have four different cases \cite{sikaroudi2020offline}. Easy and hard positives correspond to the nearest and farthest positives, respectively; easy and hard negatives correspond to the farthest and nearest negatives, respectively. Easy Positive-Easy Negative (EPEN), Easy Positive-Hard Negative (EPHN), Hard Positive-Easy Negative (HPEN), and Hard Positive-Hard Negative (HPHN) are the four possible cases. HPHN is equivalent to the batch hard method explained in Section \ref{section_batch_hard}. 
Since we are taking $k$ instances from both positive and negative sets, the cost in Eq. (\ref{equation_cost_2}) for the other three cases are as follows:
\begin{align}
&\text{$k$-EPEN:} \quad \mathcal{L} = \sum_{i,j} \eta_{ij}\,\|\b{x}_i - \b{x}_j\|^2_{\b{M}} \nonumber \\
&~~~~~~~~~~~~~~~~~~~~~~~~~~~~~ + c \sum_{i,j,l} \eta_{ij}\, \gamma_{il}^-\, (1 - y_{il})\, \xi_{ijl}, \\
&\text{$k$-EPHN:} \quad \mathcal{L} = \sum_{i,j} \eta_{ij}\, \|\b{x}_i - \b{x}_j\|^2_{\b{M}} \nonumber \\
&~~~~~~~~~~~~~~~~~~~~~~~~~~~~~ + c \sum_{i,j,l} \eta_{ij}\, \eta_{il}^-\, (1 - y_{il})\, \xi_{ijl}, \\
&\text{$k$-HPEN:} \quad \mathcal{L} = \sum_{i,j} \gamma_{ij}^+\, \|\b{x}_i - \b{x}_j\|^2_{\b{M}} \nonumber \\
&~~~~~~~~~~~~~~~~~~~~~~~~~~~~~ + c \sum_{i,j,l} \gamma_{ij}^+\, \gamma_{il}^-\, (1 - y_{il})\, \xi_{ijl},
\end{align}
where $\gamma_{il}^-$ is one (zero) if $\b{x}_l$ is (is not) amongst the $k$-farthest neighbors of $\b{x}_i$ with different class label.
The hardest cases are useful due to the concept of opposition learning \cite{tizhoosh2005opposition} and the fact that more difficult separable data points are better to be emphasized. Moreover, the easiest cases are found to be effective in the literature \cite{xuan2020improved}. 

\subsection{Negative Sampling}

In \textit{negative sampling}, as another mining method proposed for Siamese networks, for every positive instance, each negative's probability of occurrence is calculated using a stochastic probability distribution. The distribution of pairwise distances, denoted by $q(\mathcal{D})$, of two points can be estimated as \cite{wu2017sampling}:
\begin{align}
q(\mathcal{D}) \propto \mathcal{D}^{d-2}(1- 0.25 \mathcal{D}^2)^{\frac{d-3}{2}},
\end{align}
where $d$ is the dimensionality of data and $\mathcal{D}$ is defined by Eq. (\ref{equation_metric}).
For an anchor $\b{x}_i$, the probability of a negative instance $\b{x}_l$, with distance $\mathcal{D}$ from $\b{x}_i$ can be calculated as \cite{wu2017sampling}:
\begin{align}\label{equation_negative_sampling_prob}
\mathbb{P}(\b{x}_l\,|\,\b{x}_i) \propto \min(\lambda, q^{-1}(\mathcal{D})),
\end{align}
where $\lambda$ (e.g., $1.4$) is for giving all the negatives a minimum chance of selection. One can use a roulette wheel strategy for selecting negative instances using the probability in Eq. (\ref{equation_negative_sampling_prob}) \cite{talbi2009metaheuristics}. 

In this work, we select the $k$-nearest positives and sample $k$ negatives for every anchor-positive pair. We call this method \textit{$k$-negative sampling ($k$-NS)} and its cost function in Eq. (\ref{equation_cost_2}) is:
\begin{equation}
\begin{aligned}
\mathcal{L} = &\sum_{i,j} \eta_{ij}\, \|\b{x}_i - \b{x}_j\|^2_{\b{M}} + c \sum_{i,j,l} \eta_{ij}\, \rho_{il}^-\, (1 - y_{il})\, \xi_{ijl},
\end{aligned}
\end{equation}
where $\rho_{il}^-$ is one (zero) if $\b{x}_l$ is (is not) a sampled negative for the $(\b{x}_i, \b{x}_j)$ anchor-positive pair.

\section{Proposed Hierarchical Large Margin Metric Learning with Stratified Sampling} \label{section_hierarchical}

The triplet mining methods, introduced in the previous section, are promising techniques for better and faster performance of large margin metric learning; however, they can be further improved as explained here. 
We propose a hierarchical approach for accelerating the large margin metric learning. The main idea is to consider portions of data for training for solving the optimization in order to tackle the slow pace of SDP. However, for taking into account the whole training data, portions of data should be introduced to the optimization problem hierarchically. This technique has a divide and conquer manner to accelerate the training phase \cite{cormen2009introduction}. It also can improve the performance of embedding model due to model averaging \cite{claeskens2008model,breiman1996bagging} and reduction of estimation variance by stratified sampling \cite{ghojogh2019theory}. 

The procedure of this hierarchical approach can be found in Algorithm \ref{algorithm_hierarchical}. As can be seen in this algorithm, this approach is iterative. In every iteration, several hyper-spheres are considered in the space of data and the triplets are sampled from inside of the hyper-spheres (see Line \ref{alg_sampling} in Algorithm \ref{algorithm_hierarchical}). 
We employ stratified sampling \cite{barnett1974elements} where classes of data are considered to be strata.
The SDP optimization, Eq. (\ref{equation_cost_2}), is solved at every iteration using merely the sampled triplets rather than the whole data (see Line \ref{alg_optimization} in Algorithm \ref{algorithm_hierarchical}). 
We factorize the matrix $\b{M}$ in Eq. (\ref{equation_metric}) into $\b{L}\b{L}^\top$ using eigenvalue decomposition:
\begin{align}\label{equation_M_decomposition}
\b{M} = \b{\Psi} \b{\Sigma} \b{\Psi}^\top = \b{\Psi} \b{\Sigma}^{(1/2)} \b{\Sigma}^{(1/2)} \b{\Psi}^\top = \b{L} \b{L}^\top,
\end{align}
which can be done because $\b{M} \succeq 0$.
As Eq. (\ref{equation_metric}) shows, metric learning can be viewed as Euclidean distance after projection onto a subspace spanned by the columns of $\b{L}$, i.e., the column space of $\b{L}$. Hence, the whole data are projected into the metric subspace trained by the sampled triplets (see Line \ref{alg_projection} in Algorithm \ref{algorithm_hierarchical}). Note that for not having data being collapsed in subspaces with low ranks, one can slightly strengthen the diagonal of $\b{M}$ which results in larger eigenvalues without effecting the projection directions \cite{mika1999fisher}. 

At every iteration, the number of hyper-spheres, denoted by $n_s$, and the radius of them, denoted by $r$, are determined by a function decreasing and increasing with respect to the iteration index, respectively. This is because by the progress of algorithm, we want to make the hyper-spheres coarser to see more of data but at the same time, the number of them should become less not to have much overlap between the sampling areas. 
The size of stratified sampling in every hyper-sphere can also alter by the iteration index because in the late iterations, there is no need to consider the whole data in the hyper-sphere but a part of them. 
For the stratified sampling size, we sample a portion of each available class (i.e., strata) within the hyper-sphere. 
\SetAlCapSkip{0.5em}
\IncMargin{0.8em}
\begin{algorithm2e}[!t]
\DontPrintSemicolon
    \textbf{Procedure: } Hierarchical Metric Learning($\b{X}$, $k$)\;
    \textbf{Input: } $\b{X}$: dataset, $k$: number of neighbors\;
    Initialize $r$, $n_s$, and $p_\tau$\;
    \For{$\tau$ from $1$ to $T$}{
        $r := \text{increasing function of } \tau$ \;
        $n_s := \text{decreasing function of } \tau$\;
        $p_\tau := \text{decreasing function of } \tau$\;
        \For{$s$ from $1$ to $n_s$}{
            $\b{c}_s \sim \text{range}(\b{X})$\;
            $\{\b{x}_{i,a}, \b{x}_{i,p}, \b{x}_{i, n}\}_{i=1}^{n_\tau} \gets $ draw a stratified triplet sample with sampling portion $p_\tau$ within the $s$-th hypersphere\; \label{alg_sampling}
            Solve optimization (\ref{equation_cost_2})\; \label{alg_optimization}
            Decompose $\b{M}=\b{L}\b{L}^\top$ using Eq. (\ref{equation_M_decomposition})\;
            Project $\b{X}$ onto $\mathbb{C}\text{ol} (\b{L})$: $\b{X} \gets \b{L}^\top \b{X}$\; \label{alg_projection}
        }
    }
\caption{Hierarchical Large Margin Metric Learning}\label{algorithm_hierarchical}
\end{algorithm2e}
\DecMargin{0.8em}

We initialize the radius, number of hyper-spheres, and the portion of sampling by $r := 0.1 \sigma$, $n_s := \lfloor 0.01 \times n \rfloor$ (clipped to $10 \leq n_s \leq 20$), and $p_\tau := 1$. 
The updates of these variables are performed as 
$r := r + \Delta r$, $n_s := \max(n_s - \lceil 0.2 \times n_s \rceil, 1)$, and $p_\tau := \max(p_\tau - 0.05, 0.2)$, where $\Delta r := 0.3 \sigma$ and $\sigma$ is the average standard deviation along features. 

\begin{table*}[!t]
\caption{Comparing accuracies and run-time of the proposed triplet mining methods in both non-hierarchical and hierarchical metric learning for nearest neighbor classification.}
\label{table_accuracy}
\centering
\scalebox{1}{    
\begin{tabular}{l || l || c || c | c | c | c | c | c | c }
Dataset & & & $k$-BA & $k$-BH & $k$-BSH & $k$-HPEN & $k$-EPEN & $k$-EPHN & $k$-NS \\
\hline
\hline
\multirow{4}{*}{Iris} & \multirow{2}{*}{Non-Hierarchical} & Accuracy ($\%$) & 72.73 & 100 & 86.36 & 95.45 & 81.82 & 95.45 & 72.73 \\
& & Time (sec) & 832.85 & 5.51 & 6.62 & 4.77 & 5.34 & 5.11 & 5.06 \\
\cline{2-10}
& \multirow{2}{*}{Hierarchical} & Accuracy ($\%$) & 100 & 100 & 100 & 100 & 100 & 100 & 100 \\
& & Time (sec) & 23.73 & 9.72 & 4.54 & 7.25 & 4.73 & 5.05 & 4.64 \\
\hline
\multirow{4}{*}{ORL Faces} & \multirow{2}{*}{Non-Hierarchical} & Accuracy ($\%$) & -- & 85.00 & 78.75 & 72.50 & 75.00 & 85.00 & 77.50 \\
& & Time (sec) & -- & 16.13 & 18.61 & 19.59 & 19.19 & 16.31 & 19.05 \\
\cline{2-10}
& \multirow{2}{*}{Hierarchical} & Accuracy & 76.25 & 76.25 & 81.25 & 78.75 & 78.75 & 81.25 & 63.75 \\
& & Time (sec) & 0.39 & 0.93 & 0.79 & 4.36 & 1.07 & 0.95 & 0.39 \\
\hline
\multirow{4}{*}{MNIST} & \multirow{2}{*}{Non-Hierarchical} & Accuracy ($\%$) & -- & 82.00 & 79.00 & 82.00 & 78.00 & 82.00 & 78.00 \\
& & Time (sec) & -- & 122.21 & 182.13 & 152.89 & 173.18 & 135.64 & 170.33 \\
\cline{2-10}
& \multirow{2}{*}{Hierarchical} & Accuracy ($\%$) & 71.00 & 77.00 & 79.00 & 81.00 & 75.00 & 78.00 & 79.00 \\
& & Time (sec) & 27.17 & 1.56 & 1.55 & 0.49 & 1.02 & 1.70 & 1.55 
\end{tabular}%
}
\end{table*}

\begin{figure*}[!t]
\centering
\includegraphics[width=5in]{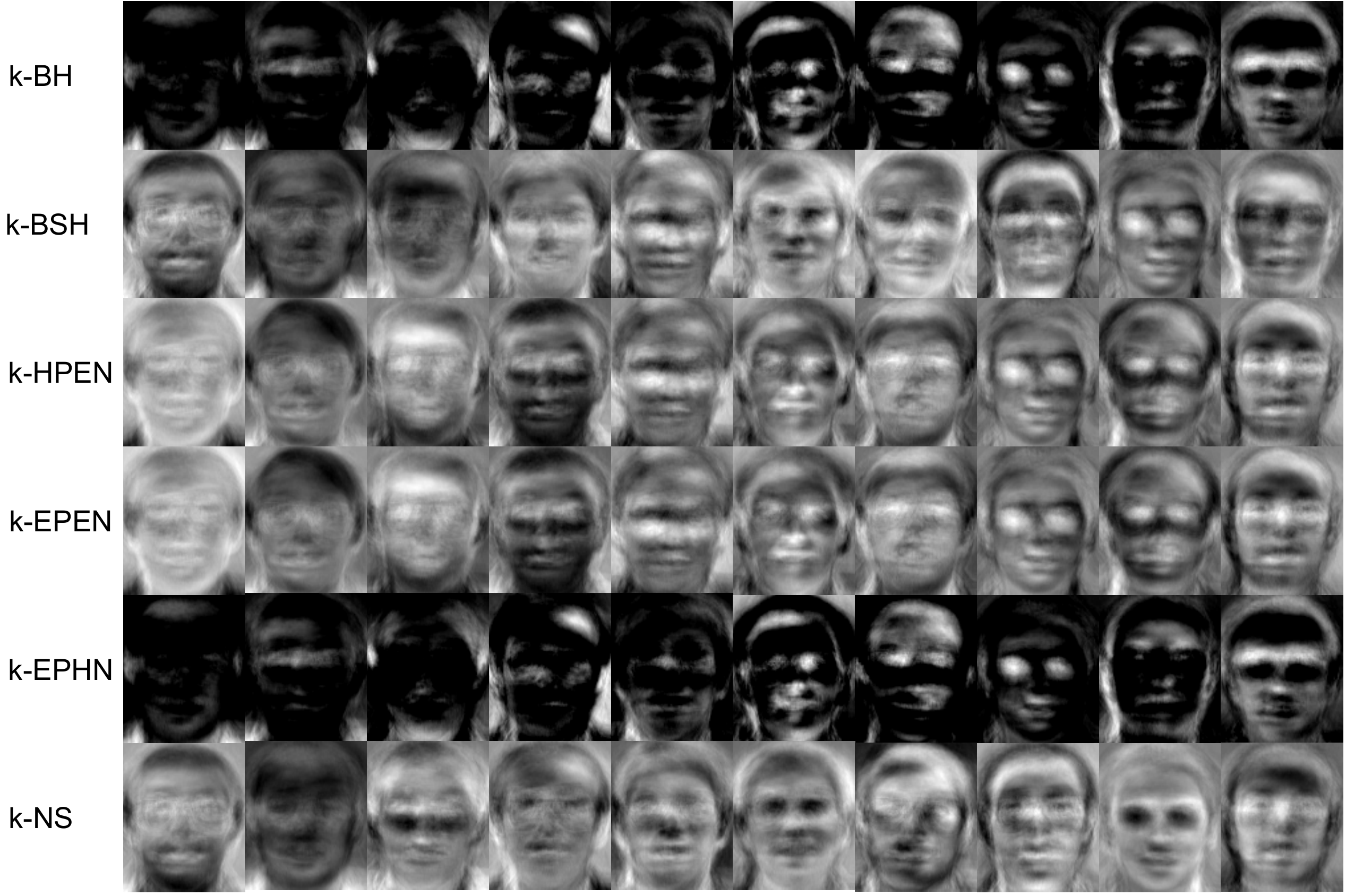}
\caption{The top ten ghost faces in different triplet mining methods.}
\label{figure_ghost_faces}
\end{figure*}

\section{Experimental Results and Analysis}\label{section_experiments}

\subsection{Datasets and Setup}

In this work, we use three publicly available datasets. The first dataset is the Fisher Iris data \cite{web_UCI_repository} which includes 150 data points in three classes with dimensionality of 4. The second dataset which we used was ORL faces data \cite{web_ORL_dataset} with 40 classes each having 10 subjects. The size of facial images are $112 \times 92$ pixels. 
The third dataset was the MNIST digits data \cite{lecun1998gradient} with $28 \times 28$-pixel images. 

The Iris dataset was randomly split into train-validation-test sets with portions $70\%$-$15\%$-$15\%$. In the ORL dataset, the first six faces of every subject made the training data and the rest of images were split to test and validation sets. A subset of MNIST with 400-100-100 images was also taken for train-validation-test.
Note that the SDP in large margin metric learning cannot handle very large datasets due to the slow pacing of optimization. 
The ORL dataset was further projected onto the 15 leading eigenfaces \cite{turk1991face} as pre-processing \cite{jolliffe2011principal}. The validation set was used for determining the optimal values of $k$ and $c$. The MNIST data were also projected onto the principal component analysis subspace with dimensionality 30. 


\subsection{Comparison of Triplet Mining Methods in the Non-Hierarchical and Hierarchical Approaches}

For each dataset, we returned the accuracy of the $k$-nearest classification using the Mahalanobis distance for the different triplet mining methods. Table \ref{table_accuracy} represents the accuracies and run-time for Iris, ORL faces, and MNIST datasets, respectively. 

In all datasets, $k$-BH has obtained the highest accuracy in non-hierarchical approach. 
However, in hierarchical approaches, $k$-BSH has obtained a top accuracy. 
The reason for $k$-BH and $k$-BSH to have acceptable performance is using the hard (near) negative instances in the training. This helps avoiding overfitting to the training data. 
In ORL faces data, the best accuracy is for $k$-BH and $k$-EPHN. This is because in both of these methods, the hardest negative instances are used for training, helping to avoid overfitting again. For the same reason, $k$-BSH has the second best performance in this dataset. Moreover, we see that the results of $k$-NS is acceptable in this data which is due to the effectiveness of the probability distribution used for sampling from the negative instances. This distribution was recently proposed for Siamese training \cite{wu2017sampling}; however, the results show that it is also effective for triplet mining in the large margin metric learning. 

In the case of Iris data, due to the small size and simplicity of dataset, the accuracies are all perfect in the hierarchical approach. In this approach, for the ORL and MNIST datasets, the highest accuracies are for $k$-BSH which can be interpreted as explained above. As obvious in table, the hierarchical approach either outperforms the non-hierarchical approach (due to model averaging) or has comparable result with much less consumed time. 

In the non-hierarchical approach, we tested the $k$-BA merely on the Iris dataset because the two other datasets are too large for $k$-BA as it considers all the negative instances. For the same reason, it is very time consuming; hence, the longest time belongs to $k$-BA in Table \ref{table_accuracy}. 
For the ORL and MNIST datasets, the longest time belongs to $k$-HPEN and $k$-BSH, respectively, mainly due to handling the hard cases in optimization. As the table shows, the hierarchical approach is scalable and much faster because of sampling. For this reason, we could run $k$-BA efficiently for all three datasets in this approach. Note that the characteristic of computer used for simulations was Intel Core-i7, 1.80GHz, with 16GB RAM. 

\subsection{Comparison of Triplet Mining Methods By Ghost Faces}

As Eq. (\ref{equation_M_decomposition}) shows, metric learning can be viewed as Euclidean distance after projection onto a subspace spanned by the columns of $\b{L}$. In the eigenvalue decomposition, the eigenvectors and eigenvalues are sorted from the leading to trailing. 

Inspired by eigenfaces \cite{turk1991face} and Fisherfaces \cite{belhumeur1997eigenfaces}, for the large margin metric learning, we can visualize the eigen-subspaces (column spaces of $\b{L}$) for the facial dataset in order to display the ghost faces. Here, we consider the top ten columns of $\b{L}$. 
The ghost faces of the ORL face dataset are depicted in Fig. \ref{figure_ghost_faces}. As seen in this figure, $k$-NS features are more discriminative which distinguish the different classes using various extracted features including eye, eyebrow, cheeks (for eye glasses), chin, hair, and nose. In second place after $k$-NS, the $k$-BSH, $k$-HPEN, and $k$-EPEN features are diverse enough (including eye, cheek, nose, and hair) for discriminating the classes. The $k$-BH and $k$-EPHN have mostly concentrated on the eye and eye-brow. This makes sense because many of the subjects in the ORL face dataset wear eye-glasses.

\section{Conclusion and Future Direction}\label{section_conclusion}

Large margin metric learning for for nearest neighbor classification makes use of SDP optimization which is very slow and computationally expensive, because of the interior point optimization method, especially when the data scale up. 
In this paper, inspired by the state-of-the-art triplet mining techniques for Siamese network training, we proposed and analyzed several triplet mining methods for large margin metric learning. These triplet mining methods make the set of triplets smaller by limiting the instances to the most important ones. This speeds up the optimization and makes it more efficient. 
The proposed triplet mining techniques were $k$-BA, $k$-BH, $k$-BSH, $k$-HPEN, $k$-EPEN, $k$-EPHN, and $k$-NS. Moreover, We suggested a new hierarchical approach which, in combination with the triplet mining methods, reduces the time of training considerably and makes the method scalable. Our experiments on three public available datasets verified the effectiveness of the proposed approaches.  
A possible future direction is to try the proposed hierarchical approach using stratified sampling on other subspace learning methods.

\bibliographystyle{IEEEtran}
\bibliography{References}

\end{document}